\documentclass[11pt]{article}
\usepackage[letterpaper,margin=1in]{geometry}
\usepackage{times}
\usepackage[round]{natbib}

\usepackage[utf8]{inputenc}
\usepackage[T1]{fontenc}
\usepackage{amsmath,amssymb,amsthm,amsfonts}
\usepackage{graphicx}
\usepackage{booktabs}
\usepackage{array}
\usepackage{tabularx}
\usepackage{microtype}
\usepackage{hyperref}
\usepackage{url}

\graphicspath{{figures/}}

\newtheorem{finding}{Finding}

\newcommand{\Wq}{\mathbf{W}_q}
\newcommand{\Wk}{\mathbf{W}_k}
\newcommand{\Wv}{\mathbf{W}_v}
\newcommand{\accor}{\mathrm{acc}^{\pm}}

\title{Probe-Geometry Alignment: Auditing and\\
Suppressing Representational Memorization\\
in Language Models}

\author{Anamika Paul Rupa\\
Department of Information Science and Systems\\
Morgan State University\\
Baltimore, MD 21251, USA\\
\texttt{anamika.rupa@morgan.edu}
\and
Anietie Andy\\
Department of Electrical Engineering\\
and Computer Science\\
Howard University\\
Washington, DC, USA\\
\texttt{anietie.andy@howard.edu}}
\date{}

\begin{document}
\maketitle

\begin{abstract}
Unlearning methods for language models are judged by what the model says. Prompt
the model with the opening words of a target passage, and if it no longer
completes the passage, the content is reported as removed. A model can pass that
test and still encode the passage. We measure what is left with a probe that has
to generalize. A linear classifier is fitted on the hidden states of several
memorized sequences, then tested on a memorized sequence it has never seen. To
succeed, it must read a property of memorization itself rather than a particular
string. It succeeds on every model we tested. Across three pretrained
architectures, the memorization-specific gap runs from $+0.19$ to $+0.35$. It is
stable across the Pythia family from 70M to 1B parameters, and it holds on a pool
whose membership is verified by corpus lookup rather than by likelihood. The
direction the probe reads is causally separable from the machinery that recites
the text. No behavioral erasure we applied removes it, from five published unlearning
objectives to weight editing to imitating a model that never memorized, however
far the output moves. That separation is what makes an efficient method possible.
Probe-geometry alignment (PGA) constrains one scalar per depth, the projection
onto the direction the probe reads, and leaves the rest of the residual stream
alone. On Pythia-70M it buys $6.6$ times as much recall suppression per nat of
capability as the best of four isotropic settings on the same recipe. It also
suppresses more recall than the behavioral baseline that retains the most
capability, while retaining more capability than that baseline, and it carries over to
Mistral-7B. Constraining a readout can also reverse it rather than empty it. Raw
probe accuracy then falls below chance while the classes stay separable, so an audit that quotes that number records erasure where none
occurred. A multi-position variant drives a fresh mean-pooled linear probe to
chance at mid depths, at a further cost in capability. The
audit and the method share one direction, so a deployment able to fit a probe
can state what its deletion claim is worth, and act on the answer.
\end{abstract}

% =============================================================================
\section{Introduction}
\label{sec:intro}

Machine unlearning for language models is evaluated by what the model says. A
method is applied, the model is prompted with the opening words of the target
text, and if it no longer completes that text the content is reported as removed.
Major frameworks, including TOFU \citep{maini2024tofu}, MUSE
\citep{shi2024muse} and WMDP \citep{li2024wmdp}, score unlearning by the model's
outputs. Membership-inference work on
pretraining data \citep{shi2024detecting} asks a related but different question,
whether a record was trained on rather than whether it is still encoded.

That criterion is not robust. Benign downstream fine-tuning brings
back content that unlearning appeared to remove \citep{hu2024jogging}, a failure
that recurs across eight evaluation axes \citep{lynch2024methods}, quantization
alone reverses the effect \citep{zhang2025catastrophic}, and suppression does not
generalize to semantically related content \citep{jia2025erasure}. Each result is
evidence that something survived, and none of them says what survived, where it
sits, or how much of it is left. Deletion is also something an organization may
have to demonstrate rather than simply perform \citep{zhang2025rtbf}, and
accountability frameworks ask that such claims rest on evidence an auditor can
examine \citep{raji2020closing}. We take the auditor to be a third party, a
reader with access to the model's weights and
activations who fits a classifier to them to test whether the target content is
still encoded.

This paper measures what remains directly. Base Pythia-70M reproduces the MIT
license preamble almost verbatim from its opening words. We write a
register-matched control, a paraphrased license of the same length and style that
the model has not memorized, and train a linear probe on hidden states to
separate the two. Any probe can do this for a single pair by reading the tokens.
The informative question is whether a probe trained on several such pairs
classifies a \emph{held-out} memorized sequence it has never seen. If it does, the hidden states carry a
feature of memorization itself rather than of any particular string. We call it
the cross-sequence memorization signature, and the leave-one-out (LOO) protocol
that measures it is the instrument this paper builds.

The instrument reports a strong and reproducible signal, positive on every model
tested and size-stable across the Pythia family at five to eight standard
deviations above a shuffled-label null. The direction it reads is causally separable
from recall: project it out of the residual stream, the running sum a transformer
carries from layer to layer, and the local signature
collapses while the model keeps reciting the text. The representation therefore
carries something the generation test does not reach, and no behavioral erasure
we applied removes it, including weight editing \citep{meng2023memit} and
distillation \citep{hinton2015distilling} from a model that never memorized.

Separating the two axes is what makes the method possible. A probe reads one
direction per depth, so an erasure objective need not move the rest of the
residual stream, most of which supports the model's other behavior. Probe-geometry alignment
constrains that one scalar. This is where its efficiency comes from, and also
where a measurement problem appears, because constraining what a probe reads can
reverse that reading rather than remove it.

The paper makes four contributions.

\begin{enumerate}
\item \textbf{A protocol that makes representational memorization measurable}
(Sections~\ref{sec:protocol}, \ref{sec:signature} and~\ref{sec:separable}). The
cross-sequence LOO probe asks whether a memorization signature
\emph{generalizes} to held-out sequences. It needs neither the training corpus
nor a pre-unlearning reference model \citep{xu2025unlearning}, and it tests a
specific passage rather than whether a model was unlearned at all
\citep{chen2025invisible}. The signature is positive on all three architectures
we test, and the direction it reads is causally separable from recall:
projecting it out collapses the local signature from $+0.643$ to $+0.161$ while
recall falls by at most $0.116$ nats per token.

\item \textbf{Evidence that behavioral erasure does not reach the encoding}
(Section~\ref{sec:existing}). Five published unlearning methods leave the probe
at $1.000$ on a controlled testbed, a small transformer trained to recall one
injected sentence, and four show the same pattern on Pythia-70M while changing
the output by up to $62$ nats. Weight editing leaves the probe at $1.000$, and
distillation from a teacher that never memorized erases the behavior at almost
no capability cost yet leaves the probe at $0.922$.

\item \textbf{Probe-geometry alignment (PGA), which constrains only what the
probe reads} (Sections~\ref{sec:pga}, \ref{sec:isotropic}
and~\ref{sec:attackers}). PGA constrains one scalar per depth. On Pythia-70M it
suppresses more recall than refusal tuning (IDK), the baseline that keeps the
most capability,
while also keeping more capability ($3.25$ against $1.05$ nats, at $88.6\%$
against $79.4\%$ retention). An isotropic penalty on the same recipe, layers and
budget is $6.6$ times less efficient at its best setting and $19.5$ times less
at PGA's own penalty weight, so the gain comes from the geometry rather than
from penalty strength. PGA also suppresses recall on Mistral-7B, and a
multi-position variant drives a fresh mean-pooled linear probe to chance at mid
depths.

\item \textbf{Raw probe accuracy is not a safe erasure criterion}
(Section~\ref{sec:existing} and Appendix~\ref{app:orientation}). Alignment can
reverse a readout rather than remove it. A probe fitted afterwards then gets
almost every label wrong, so its raw accuracy falls toward $0$ rather than
toward chance, and an auditor quoting it would record erasure where the classes
remain separable. Raw accuracy falls below chance at all four scales we test,
and of ten configurations audited both ways only the three trained PGA settings
change verdict. Reporting orientation-corrected accuracy, $\max(\mathrm{acc},
1-\mathrm{acc})$ \citep{burns2023discovering}, avoids the error, and the point applies to any erasure claim
audited with a probe.
\end{enumerate}

\begin{figure}[t]
\centering
\includegraphics[width=\linewidth]{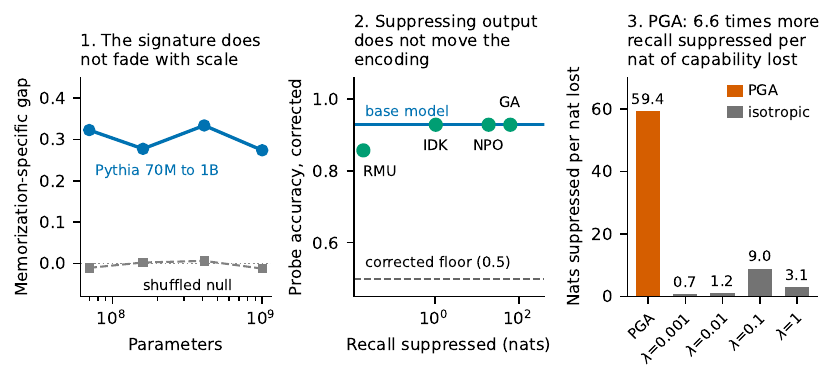}
\caption{Summary of the three results.}
\label{fig:overview}
\end{figure}

Figure~\ref{fig:overview} summarizes the evidence in three panels: the signature holds across
scale, four existing methods move the output and not the signature, and aligning
the probe's own direction suppresses far more recall per nat of capability than
acting on the whole residual stream. The four are gradient ascent (GA), negative preference
optimization (NPO), representation misdirection for unlearning (RMU) and IDK.

% =============================================================================
\section{Related work}
\label{sec:related}

Machine unlearning removes the
influence of specific training data without retraining from scratch
\citep{cao2015towards,bourtoule2021machine}. Recipes range from
Fisher-information weight scrubbing \citep{golatkar2020eternal} and surgical
replacement \citep{eldan2023whos} to NPO
\citep{zhang2024negative} and gradient ascent on negative examples
\citep{yao2023large}, dozens of which \citet{ren2025sok} catalog. What unites
them is the evaluation, and by that measure most succeed. The robustness failures
cited in Section~\ref{sec:intro} infer retention indirectly, as do \citet{chen2025does} and
\citet{wang2025ilu}, who establish it by exhibiting a recovery procedure. We
measure the retained structure directly, at a specified depth against a stated
null, which an auditor can run without a recovery procedure in hand.

A linear probe that separates a class lower-bounds what the hidden state carries
\citep{marks2023geometry,hewitt2019structural}, provided its accuracy is read
against a null rather than on its own \citep{belinkov2022probing}. A probe that
still classifies after unlearning is therefore evidence the structure survived.
Knowledge editing localizes factual recall to mid-layer feed-forward modules,
where a targeted weight edit changes the output
\citep{meng2022locating,meng2023memit}, but an intervention that succeeds at one
site does not establish that the information was stored only there. Activation
steering makes the point from the other direction, shifting behavior with a
vector added to the residual stream and no weights touched
\citep{turner2023steering}. We use causal tracing \citep{meng2022locating} to
identify the minimal circuit \citep{elhage2021mathematical} behind a memorized
completion, and the probe to test whether an intervention reached storage or only
silenced output.

What this paper measures is a property that transfers between sequences, which
is what the question demands when it asks whether \emph{this passage} is still
encoded rather than whether \emph{this model} was tampered with. It is also what
lets one instrument serve both as an audit and as the target of a suppression
objective. Related diagnostics read latent knowledge item by item
\citep{lynch2024methods}, quantify drift against a pre-unlearning copy
\citep{xu2025unlearning}, or train a classifier to detect whether a model has been
unlearned at all \citep{chen2025invisible}. None tests whether the signature
generalizes to held-out sequences, and none turns its diagnostic into a training
objective. Table~\ref{tab:prior} sets out the comparison.

Concept erasure supplies the closest methodological neighbors, each removing a
concept that a linear classifier could read: the nullspace projection of INLP
\citep{ravfogel2020null}, the minimax of R-LACE \citep{ravfogel2022rlace}, and
the closed-form guarantee of LEACE \citep{belrose2023leace}. Those guarantees
hold when the concept is fixed and known in advance. Memorization is neither,
which is why the method of Section~\ref{sec:pga} refits its direction during
training rather than solving once.

\begin{table}[t]
\centering
\small
\caption{Comparison with prior work.}
\label{tab:prior}
\begin{tabularx}{\textwidth}{@{}l>{\raggedright\arraybackslash}X c c c@{}}
\toprule
Study & Measurement & Measurement needs & Cross-sequence & Suppression \\
 & & reference model & transfer & method \\
\midrule
\citet{hu2024jogging} & Recovery after fine-tuning & No & No & No \\
\citet{lynch2024methods} & Latent-knowledge probe & No & No & No \\
\citet{zhang2025catastrophic} & Recovery after quantization & No & No & Yes \\
\citet{jia2025erasure} & Recall on related content & No & No & No \\
\citet{chen2025does} & Prompt and activation audits & No & No & No \\
\citet{chen2025invisible} & Unlearning-trace classifier & No & No & No \\
\citet{xu2025unlearning} & Representational drift & Yes & No & No \\
\citet{belrose2023leace} & Fixed-concept recoverability & No & No & Yes \\
\citet{zhang2024negative} & Behavioral recall (NPO) & No & No & Yes \\
\citet{maini2024tofu} & Match to retrain (IDK) & Yes & No & Yes \\
\citet{li2024wmdp} & Hazard question accuracy & No & No & Yes \\
\citet{wang2025ilu} & Recall after later fine-tuning & No & No & Yes \\
This work & Held-out probe transfer & No & Yes & Yes \\
\bottomrule
\end{tabularx}
\end{table}

% =============================================================================
\section{Methods}
\label{sec:methods}

\subsection{The cross-sequence leave-one-out protocol}
\label{sec:protocol}

A probe trained to separate one memorized string from one control reaches
accuracy $1.000$ on every model we tested, and learns the tokens rather than
anything about memorization. The LOO design removes that shortcut by requiring
the probe to classify a sequence it has never seen.

For each model we build three per-sequence pools on shared identifiers, one key
per source text such as \texttt{mit\_license}.
$\mathrm{POOL}_M$ holds sequences passing a per-model screen on mean
log-probability per token. $\mathrm{POOL}_C$ holds a register-matched paraphrase
per key, built by substituting distinctive terms so that register and surface
form are preserved while the model no longer reproduces the text. Where
$\mathrm{POOL}_M[\texttt{mit\_license}]$ begins ``Permission is hereby granted'',
$\mathrm{POOL}_C[\texttt{mit\_license}]$ begins ``Permission is explicitly
denied, under all conditions''. $\mathrm{POOL}_D$ holds a second, further
paraphrased variant per key. A shared pool $\mathcal{B}$ of $24$ neutral prefixes
is prepended at extraction time to supply within-class variance. Thresholds are
in Appendix~\ref{app:protocol} and the verbatim pools in the released archive.

At each residual-stream depth $d$, with $d=0$ the embedding output and one
depth per transformer layer thereafter, and for each held-out key, we fit a
logistic regression probe on the activations of all remaining keys, labeling $\mathrm{POOL}_M$ positive and $\mathrm{POOL}_C$
negative, and evaluate on the held-out key's $24$ memorized and $24$ control
variants.
Averaging over held-out keys gives the \emph{true LOO} accuracy. Two nulls
calibrate it. The \emph{shuffled-label} null permutes training labels within each fold
\citep{hewitt2019designing}.
The \emph{pure-distinguishability} null repeats the procedure with
$\mathrm{POOL}_C$ positive and $\mathrm{POOL}_D$ negative, both screened as
non-memorized, and measures how separable two register-matched paraphrases are
from one another. That second null matters most, because it measures the
nuisance separability a naive reading would mistake for memorization. The
memorization-specific gap is their difference,
\begin{equation}
g_d \;=\; \mathrm{LOO}_{\mathrm{true}}(d) \;-\; \mathrm{LOO}_{\mathrm{pure}}(d).
\label{eq:gap}
\end{equation}
We summarize a model by the mean of $g_d$ over transformer depths, excluding the
embedding output, where token identity is trivially readable. Throughout, the
\emph{base model} is the model before any unlearning or alignment.
\emph{Recall suppression} is its mean $\log P/\mathrm{tok}$ on the memorized pool
minus the same quantity after an intervention, in nats per token, the natural-logarithm unit of log-probability.
\emph{Held-out cost} is the corresponding drop in mean
$\log P/\mathrm{tok}$ on held-out text, also per token, so the ratio of the two is
dimensionless.

A linear readout is defined only up to sign, since negating the weight vector
costs nothing and reverses every prediction. Where the question is what an adversary can
recover, we report accuracy orientation-corrected, $\accor = \max(\mathrm{acc},
1-\mathrm{acc})$, the convention \citet{burns2023discovering} use for a probe
whose sign is not fixed. Because the correction folds every reading below $0.5$ upward, an
uninformative readout scores above $0.5$ in expectation. Wherever we ask whether a
corrected accuracy is above chance, we compare it against a permutation null of
$200$ label permutations \citep{ojala2010permutation} rather than against $0.5$, and report the null's $95$th
percentile. For the statistic used in Sections~\ref{sec:existing}
to~\ref{sec:attackers}, the maximum corrected logistic-regression accuracy over
depths two to six on Pythia-70M, that percentile is $0.929$. Eq.~\eqref{eq:gap} subtracts raw accuracies, since correcting a
chance-centered null would inflate it (Appendix~\ref{app:protocol}).

% =============================================================================
\subsection{Probe-geometry alignment}
\label{sec:pga}

A suppression method has to move the readout of Section~\ref{sec:protocol}, and
this section defines the one we evaluate. To make memorized and control
activations look alike, the natural move acts on the full
$d_{\mathrm{model}}$-dimensional difference between them, whether by penalizing
it directly as in FitNets \citep{romero2015fitnets}, through a minimax game
against a discriminator \citep{ganin2015unsupervised}, or by the concept-erasure
constructions of Section~\ref{sec:related}. Most of that difference supports the model's other behavior, so constraining all
of it costs capability in directions the fitted probe does not read. We therefore compare against an isotropic penalty run on the
same recipe (Section~\ref{sec:isotropic}) rather than against LEACE, whose
guarantee assumes the concept it erases is fixed.

Probe-geometry alignment (PGA) constrains only the direction the auditor reads.
At each
depth $d$ we hold a unit-norm readout direction $\hat w_d$, refit from the current
model every $K$ steps, $25$ on Pythia-70M, and penalize only the scalar projection of the
memorized-against-control difference onto it:
\begin{equation}
L_{\mathrm{PGA}}(\theta) =
\underbrace{\lambda_c\,\mathrm{CE}_{\mathrm{clean}}(M_\theta)}_{\text{capability anchor}}
+ \frac{\lambda}{N\lvert\mathcal{A}\rvert} \sum_{d \in \mathcal{A}} \sum_{i=1}^{N}
\Bigl( \hat w_d^{\top} \bigl[
h_{d,\theta}(P^{\mathrm{sec}}_i) - h_{d,\theta}(P^{\mathrm{cln}}_i)
\bigr]_{\mathrm{pos}=p_{\mathrm{len}}-1} \Bigr)^{2}.
\label{eq:pga}
\end{equation}
Here $\mathcal{A}$ is the set of aligned depths, which excludes the embedding
output on the pretrained models, $N$ is the number of memorized and control pairs, and $\lambda_c$ is $1$
in every run except Mistral-7B, where the anchor is weighted at $3$.
Appendix~\ref{app:configs} gives $\mathcal{A}$ and $\lambda$ for every run, and
Appendix~\ref{app:scope} the claim each supports. Each
depth in $\mathcal{A}$ contributes one scalar constraint rather than
$d_{\mathrm{model}}$ of them, and the direction is refit because it moves under
the edit. If the penalty term goes to
zero, every pair's projected difference along $\hat w_d$ goes with it, and so
does the difference of class means along that direction. That removes the
first-moment signal a readout along $\hat w_d$ uses, and under equal
class-conditional covariances it bounds accuracy along that one direction by the
majority-class baseline. The bound covers the constrained direction, not a probe
free to choose another. Sections~\ref{sec:isotropic} and~\ref{sec:attackers} measure what
the objective achieves against an attacker given that freedom.

Equation~\eqref{eq:pga} is the training objective, and every result labeled
PGA uses it unless stated otherwise. The same geometry also works without
training. Inference PGA removes $k$ probe directions from the residual stream at
inference, refitting the probe after each removal, and the zero-shot benchmarks
of Section~\ref{sec:capability} use it with $k=1$ and with $k=4$, the number of
directions needed to bring the probe below $0.55$. On GPT-2 Medium the pool
supplies far fewer examples than $d_{\mathrm{model}}$, which leaves a single
fitted direction under-determined, so rank-$k$ PGA (RK-PGA) projects out, also at
inference, the $k$-dimensional subspace spanned by the top-$k$ eigenvectors of the
standardized between-class scatter and variance-difference matrix, with $k=2$.
Multi-position PGA, defined in Section~\ref{sec:attackers} where it is used, moves
the trained constraint from the final token to mean-pooled representations. The
Pythia-70M and Mistral-7B runs train a low-rank adaptation (LoRA) adapter
\citep{hu2022lora}, and the controlled testbed trains full weights.

% =============================================================================
\section{Results}
\label{sec:results}

\subsection{The cross-sequence memorization signature}
\label{sec:signature}

\begin{figure}[t]
\centering
\includegraphics[width=0.86\linewidth]{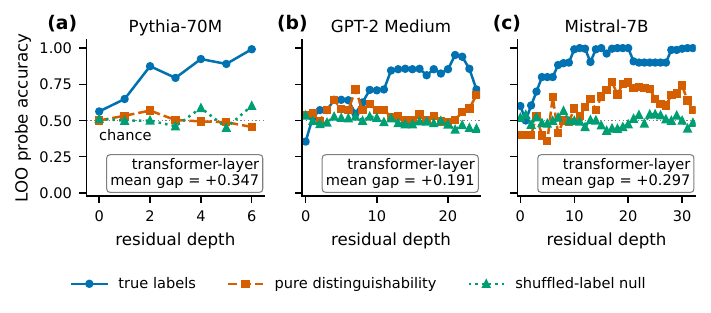}
\caption{Cross-sequence signature across three architectures.}
\label{fig:loo}
\end{figure}

The probe generalizes in all three architectures, from the middle layers onward.
Figure~\ref{fig:loo} plots true LOO accuracy against residual depth for
Pythia-70M \citep{biderman2023pythia,black2022gptneox,gao2020pile},
GPT-2 Medium \citep{radford2019language} and Mistral-7B \citep{jiang2023mistral}
against both nulls, with the transformer-layer mean of Eq.~\eqref{eq:gap}
annotated in each panel. The shuffled-label null sits near $0.5$ throughout. The
pure-distinguishability null is higher and structured, reaching roughly $0.7$ on
GPT-2 Medium and $0.8$ on Mistral-7B in mid network, which is the nuisance a
study without this control would score as memorization. The true curve clears
both.

Transformer-layer mean gaps are $+0.347$ on Pythia-70M, $+0.191$ on GPT-2 Medium
and $+0.297$ on Mistral-7B, peaking at $+0.537$, $+0.449$ and $+0.471$. A
$100$-replicate bootstrap over neutral contexts on GPT-2 Medium gives a $95\%$
interval of $[+0.184, +0.203]$ around that run's own point estimate of $+0.195$. A seven-fold sequence jackknife leaves every
held-out gap strictly positive, in $[+0.095, +0.250]$.

Running the identical protocol with the pool fixed at seven sequences on Pythia
70M, 160M, 410M and 1B gives transformer-layer means of $+0.323$, $+0.278$,
$+0.334$ and $+0.274$, which sit $8.0$, $6.5$, $8.2$ and $5.1$ standard
deviations above each model's own shuffled-label null. The peak migrates in relative
depth, so the signature is not pinned to a fixed architectural position
(Appendix~\ref{app:scaling}).

Four controls test the obvious alternative explanations, for architecture,
register, vocabulary and content type. A randomly initialized Pythia-70M shows
that architecture alone carries part of the effect and pretraining supplies the
rest: its transformer-layer mean gap is $+0.188$ against the pretrained
$+0.323$, and the two profiles run in opposite directions, random initialization
peaking at $+0.327$ in the first layer and falling to $-0.042$ in the last, while
the pretrained model peaks at $+0.643$ at depth four.
On GPT-2 Medium the
probe reads held-out sequences from all four content clusters above the
pure-distinguishability null, so no single register carries the result
(Appendix~\ref{app:controls}).

Those pools are selected by likelihood screening. Corpus lookup on the Pile
\citep{gao2020pile} instead lifts the pool to $30$ sequences and makes the
controls constructed rather than authored. The corrected gap peaks at $+0.413$,
where the probe's accuracy sits $10.5$ standard deviations above a permutation
null, with a thirty-fold jackknife
spanning only $[+0.362, +0.459]$ (Appendix~\ref{app:verified}).

\begin{finding}
\label{find:signature}
A memorization-specific signature generalizes to held-out sequences in every
model tested, from a 70M-parameter GPT-NeoX to a 7.24B-parameter Mistral, and is
size-stable within a family from 70M to 1B at $+0.27$ to $+0.33$, five to eight
standard deviations above a shuffled-label null. On a verified-membership pool of
$30$ Pile sequences the corrected gap reaches $+0.413$, where the probe's accuracy
sits $10.5$ standard deviations above a permutation null.
\end{finding}

% =============================================================================
\subsection{Causal separability of the probe direction}
\label{sec:separable}

If the direction the probe reads were the direction producing the memorized
completion, suppressing it would amount to unlearning by another name. The two
come apart. On Pythia-70M, seven-sequence pool, we fit the LOO probe at the peak-gap depth four
and intervene at that depth to
project its unit-norm weight out of the residual stream, $h' = h - (h
\cdot \hat w)\hat w$. The gap collapses at that depth, from $+0.643$ to
$+0.161$, partly persists at depth five and partly recovers by depth six, with the
transformer-layer mean falling from $+0.323$ to $+0.193$
(Appendix~\ref{app:probedir}). The model continues to recite the text.
Memorized $\log
P/\mathrm{tok}$ falls by between $0.053$ and $0.116$ nats across the seven
sequences, and held-out log-probability moves from $-5.138$ to $-5.163$, a
degradation of $0.025$ nats.

Downstream layers partially reconstruct the signature, so the structure is
re-derived as computation proceeds. That is why a projection at a single depth
does not hold and why Section~\ref{sec:pga} constrains a set of depths during
training.

\begin{finding}
\label{find:separable}
Probe-readable structure and recall-producing structure occupy separable
directions. Projecting out the fitted probe direction collapses the local
signature while behavioral recall drops by at most $0.116$ nats per token.
\end{finding}

% =============================================================================
\subsection{Existing unlearning methods}
\label{sec:existing}

Whether existing unlearning methods reach this structure is cleanest to test on
a controlled model. On a four-layer
transformer \citep{vaswani2017attention} trained to recall an injected sentence
with probability $1.000$, causal tracing over attention heads
\citep{meng2022locating,wang2023interpretability} localizes the behavior to one
head out of $32$, and an edit restricted to that
head drives recall to $1.3 \times 10^{-4}$ while touching $0.76\%$ of the
parameters. This is the best case for precision unlearning, since the circuit is
localized to one head and the edit is confined to it, and the probe stays at $1.000$. Five
published methods spanning both families of \citet{ren2025sok} behave the same
way on that testbed: four reach behavioral erasure, gradient ascent does not, and
all five leave the probe at $1.000$. Appendix~\ref{app:toy} gives the
per-method numbers, and Appendices~\ref{app:multisecret}, \ref{app:clpa}
and~\ref{app:relearning} extend the result to eight independently trained
secrets, to alignment restricted to the causally identified heads, and to
relearning.

Two stronger tests on a nine-secret version of the testbed point the same way
(Appendix~\ref{app:stronger}).
Mass-editing memory in a transformer (MEMIT) \citep{meng2023memit} edits the
weights that hold the association rather than training against the output, and
takes recall from $0.978$ to $8.4 \times 10^{-5}$ with the probe still at
$1.000$. Distillation from a teacher that never memorized comes closest to
retraining, since it has a model to imitate rather than an
objective to descend. It drives the student's recall to $2.6 \times 10^{-5}$ at
a held-out perplexity of $1.405$ against $1.403$, and the student still reads
$0.922$ where the teacher's best depth reads $0.661$. A model can be trained to
reproduce the behavior of one that never memorized and keep a geometry that one
does not have.

\begin{figure}[t]
\centering
\includegraphics[width=0.9\linewidth]{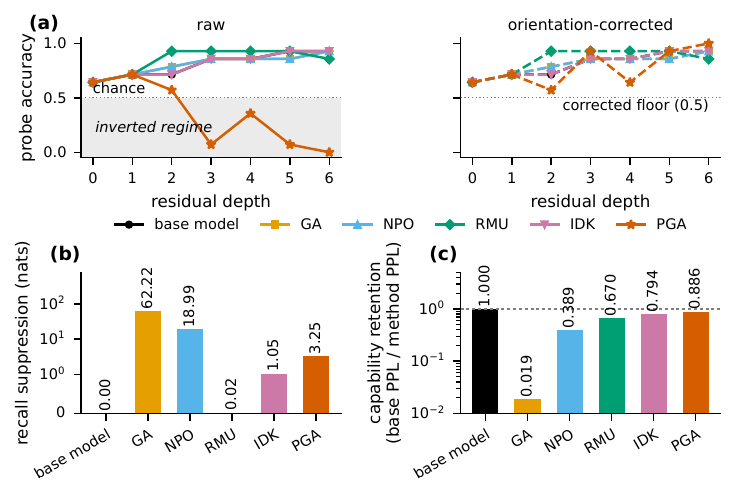}
\caption{Five unlearning methods on Pythia-70M.}
\label{fig:baselines}
\end{figure}

The same pattern holds at pretrained scale. Table~\ref{tab:baselines} and
Figure~\ref{fig:baselines} compare PGA on Pythia-70M, with the natural
memorized pool and one scoring protocol, against GA \citep{yao2023large}, NPO
\citep{zhang2024negative}, RMU \citep{li2024wmdp} and IDK \citep{maini2024tofu}. The
figure gives probe accuracy by depth under both conventions, recall suppression
in nats, and capability retention, and only PGA drives the raw reading below chance.
SimNPO \citep{fan2026simplicity} is run on the controlled testbed only.

\begin{table}[t]
\centering
\small
\caption{Unlearning methods compared on Pythia-70M, unified pipeline. PPL is
held-out perplexity, retention is base-model PPL over method PPL, and the probe
column is orientation-corrected accuracy at the deepest depth.}
\label{tab:baselines}
\begin{tabular}{@{}lrrrr@{}}
\toprule
Method & Recall supp. (nats) & Held-out PPL & Capability retention & Probe at depth 6 \\
\midrule
Base model & $0.00$   & $191.1$   & $1.000$ & $0.929$ \\
GA         & $62.22$  & $10198.5$ & $0.019$ & $0.929$ \\
NPO        & $18.99$  & $491.1$   & $0.389$ & $0.929$ \\
RMU        & $0.02$   & $285.3$   & $0.670$ & $0.857$ \\
IDK        & $1.05$   & $240.6$   & $0.794$ & $0.929$ \\
PGA (ours) & $3.25$   & $215.5$   & $0.886$ & $1.000$ \\
\bottomrule
\end{tabular}
\end{table}

The probe column does not separate these methods: every one, ours included,
leaves the deepest-depth readout between $0.857$ and $1.000$ while recall
suppression ranges over three orders of magnitude. The controlled testbed showed
the same dissociation, now reproduced with natural memorized content at
pretrained scale.

What the other two columns separate is the price. RMU, the only baseline designed
to redirect internal representations, delivers $0.02$ nats of suppression at a
cost of a third of the model's capability. IDK delivers $1.05$ nats at a cost of
a fifth. GA and NPO suppress far more by destroying the model, GA at $1.9\%$
retention. PGA delivers $3.25$ nats at $88.6\%$ retention, which
Section~\ref{sec:isotropic} examines.

The $1.000$ in the PGA row needs reading carefully, because the raw number behind it
is the one an
audit would quote. The probe column is a fresh leave-one-out logistic regression
fit on the aligned model at each depth. At the deepest depth it places every
held-out sequence on the wrong side, so the raw reading is $0.000$ and an audit
quoting it would certify erasure. Alignment did not remove the class information.
It reversed the orientation of the readout, and the corrected value says that a
reader who fixes the sign recovers the classes in full. The two readings are
complements of one another, and only the corrected one describes what an
adversary can do, since the sign flip is free. A stronger audit gives the
attacker eight classifier families rather than one, and under the protocol of
Section~\ref{sec:isotropic} the best of the eight reads $0.896$ against $0.955$
for the base model.

The six rows come from a single pipeline that scores every method with the same
probe. The leave-one-out protocol of
Section~\ref{sec:protocol} reads the same PGA model at $0.857$ at the deepest depth, and $0.893 \pm 0.036$ across four
adapter seeds (Appendix~\ref{app:configs}).

% =============================================================================
\subsection{An isotropic ablation}
\label{sec:isotropic}

Table~\ref{tab:isotropic} measures what the geometric targeting buys by
replacing only the geometry. It holds the recipe, layers, seeds, budget and
held-out estimator fixed and substitutes an isotropic alignment, a
$d_{\mathrm{model}}$-dimensional squared-distance penalty swept over $\lambda \in \{0.001, 0.01, 0.1, 1.0\}$, the last matching
PGA's own weight.

\begin{table}[t]
\centering
\small
\caption{Suppression efficiency under a common estimator, Pythia-70M.}
\label{tab:isotropic}
\begin{tabular}{@{}lrrr@{}}
\toprule
Variant & Recall supp. (nats) & Held-out cost (nats) & Nats suppressed per nat lost \\
\midrule
PGA, $\lambda = 1.0$          & $3.25$ & $0.055$ & $\mathbf{59.40}$ \\
Isotropic, $\lambda = 0.001$  & $4.27$ & $6.107$ & $0.70$ \\
Isotropic, $\lambda = 0.01$   & $5.58$ & $4.665$ & $1.20$ \\
Isotropic, $\lambda = 0.1$    & $4.76$ & $0.532$ & $8.96$ \\
Isotropic, $\lambda = 1.0$    & $5.44$ & $1.781$ & $3.05$ \\
\bottomrule
\end{tabular}
\end{table}

The isotropic ablation suppresses recall effectively but not efficiently. It
returns $8.96$ nats per nat of held-out capability at its most efficient
setting and $0.70$ at its least. PGA returns $59.4$, a factor of $6.6$ over the
best isotropic setting and $19.5$ over the arm trained at the same penalty
weight. Against the eight-family probe suite of Section~\ref{sec:attackers}, read at the
last token that both methods align, PGA's highest corrected readout is $0.896$, below the ablation's
$0.935$ and the base model's $0.955$, so the directional objective leads on both
efficiency and probe suppression.

The margin holds against the best of the four weights tested and against the arm
trained at PGA's own weight, which removes the penalty budget as an explanation.
With the held-out estimator of the unified pipeline instead, PGA's ratio is
$26.9$ (Appendix~\ref{app:margin}).

\subsection{Adaptive attackers and multi-position alignment}
\label{sec:attackers}

PGA is optimized against a probe family, so we evaluate it against attackers from
outside that family. We hand an attacker eight classifier families spanning linear,
kernel, instance-based and ensemble methods, free choice of regularization and
seed, free negation of the fitted weights, and a pooling scheme PGA never
constrains: PGA aligns the last token, and the attacker mean-pools over all
positions.

Against last-token PGA on Pythia-70M, a fresh mean-pooled logistic
regression reads $0.914$, $0.940$ and $0.955$ at depths three to five, and the
strongest of the eight families reaches $0.979$. The alignment did not generalize
across the position axis, a specific diagnosis that points at its own repair.

Multi-position PGA (MP-PGA) extends the alignment to mean-pooled representations,
identical in every other respect. The fresh mean-pooled probe falls to $0.548$ at
depth three, below the chance expectation of the corrected metric
(Appendix~\ref{app:protocol}), while recall suppression \emph{rises} from $3.25$
to $4.84$ nats, at $0.324$ nats more held-out cost than last-token
alignment. The result is scoped to the linear attacker at
mid depths. The strongest non-linear family there still reads $0.789$ and the
deepest depth returns to $0.929$, which is where the construction extends next
(Appendix~\ref{app:mp}, Figure~\ref{fig:mp}).

\subsection{Capability retention and scale}
\label{sec:capability}

Removing the probe direction costs little general capability. Inference PGA
with $k=1$ and with $k=4$ moves no zero-shot benchmark by more than $2.9$ percentage points, with mean changes of $-0.52$ and
$+0.19$ points across five tasks. At Mistral-7B scale, PGA suppresses recall by $2.63$ nats
at a held-out cost of $0.72$ nats, which is $3.6$ nats per nat. The readout moves with it: corrected accuracy at the five
aligned depths falls from $0.988$--$1.000$ to $0.500$--$0.714$, while the
best-reading depth, which alignment never targeted, still reads $0.881$. Appendices~\ref{app:capability} and~\ref{app:scope} give the per-task numbers
and the scale and configuration behind each claim.

% =============================================================================
\section{Discussion}
\label{sec:discussion}

This paper turns representational memorization into something an auditor can
measure. One probe, fitted on a handful of memorized texts, reports whether a
passage it has never seen is still encoded. The signature holds on every model we tested and is causally separable from the machinery that recites the
text, which is why a behavioral test cannot see it and why no behavioral
erasure we applied reaches it, from five published unlearning objectives to weight editing
to imitation of a model that never memorized.

The same measurement doubles as a training target, and that is where the
efficiency comes from. Probe-geometry alignment constrains one scalar per depth,
leaves the rest of the residual stream alone, and returns $6.6$ times the recall
suppression per nat of capability of the best isotropic setting. One result applies beyond this method: a constrained readout can turn
over rather than empty, and a reversed readout
scores below chance and resembles success, so any erasure claim measured with a
probe has to state which way the probe points. Appendix~\ref{app:orientation}
reports both conventions for ten testbed configurations and at all four scales.

\paragraph{Limitations and next steps.} The instrument measures what an auditor
can compute from weights and activations alone, which is whether the target
content is still linearly decodable. Three extensions follow directly from that.
Establishing pool membership by corpus lookup, as we do for Pythia-70M, applies
to any model whose training corpus is public. The protocol needs only new pools
to run on content types beyond the software licenses and web boilerplate used
here, and on models larger than the 7B we reach. Pairing the audit with
extraction measurements would turn a statement about encoding into a risk
estimate.

The instrument and the method share one object. What an auditor computes from
weights and activations alone is exactly what alignment constrains, so any
deployment able to fit a probe can state what its deletion claim is worth, and
act on the answer.

% =============================================================================
\bibliographystyle{plainnat}
\bibliography{references}

\newpage
\appendix

\section{Multi-position alignment}
\label{app:mp}

Last-token PGA constrains the readout at one position. The attacker of
Section~\ref{sec:attackers} pools over positions, and that is the axis along
which the alignment failed to generalize. Multi-position PGA extends the same
one-scalar-per-depth constraint to mean-pooled representations, leaving the
loss weights, the refit schedule, the layers, the seeds and the capability anchor
unchanged.

The repair is specific to the axis the diagnosis named. Against a fresh
mean-pooled logistic regression the readout falls to $0.548$, $0.568$ and $0.548$
at depths three to five, below the chance expectation of the corrected metric, while recall suppression rises from $3.25$ to $4.84$ nats. Because the
objective and the attacker now share a pooling scheme, this is confirmation that
the account localizes the failure correctly rather than an independent
prediction. The strongest non-linear family at those depths reads $0.789$, $0.732$ and
$0.702$, the two multilayer perceptrons fall with the linear probes, and the
final depth returns to $0.929$.

\begin{figure}[t]
\centering
\includegraphics[width=\linewidth]{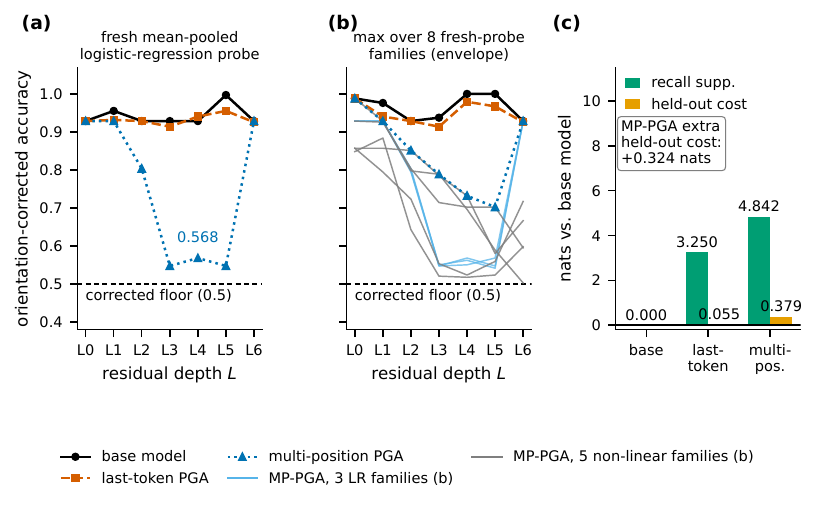}
\caption{Multi-position alignment against a mean-pooled attacker. (a) A fresh
mean-pooled logistic regression. (b) The maximum over eight fresh-probe families.
(c) Recall suppression and held-out cost for both variants.}
\label{fig:mp}
\end{figure}

\section{Probe-direction ablation}
\label{app:probedir}

\begin{figure}[h]
\centering
\includegraphics[width=\linewidth]{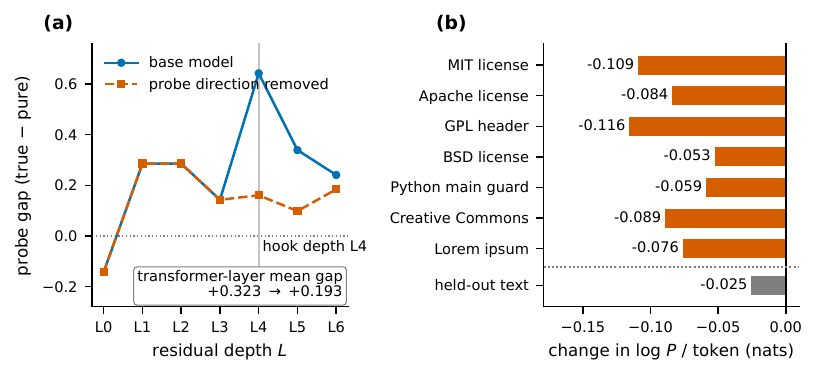}
\caption{Ablating the probe direction on Pythia-70M.}
\label{fig:probedir}
\end{figure}

Figure~\ref{fig:probedir} shows the intervention of
Section~\ref{sec:separable}. Panel (a) gives the memorization-specific gap by
depth before and after the fitted probe direction is projected out of the
residual stream at depth four. Panel (b) gives the change in $\log P/\mathrm{tok}$
that the same intervention causes for each memorized sequence and for held-out
text. No memorized sequence moves by more than $0.116$ nats.

\section{Leave-one-out protocol details}
\label{app:protocol}

The memorization-specific gap of Eq.~\eqref{eq:gap} subtracts raw leave-one-out
accuracies, while the readout numbers reported in Section~\ref{sec:results} are
orientation-corrected. The two conventions answer different
questions. The corrected metric asks what an adversary can recover, and the sign
flip is free to an adversary. A null, by contrast, is centered on chance, and
$\max(a, 1-a)$ applied to a chance-centered statistic has expectation strictly
above $0.5$: simulating an uninformative probe gives a corrected mean of $0.656$
at seven folds and $0.572$ at thirty, against a raw mean of $0.500$ in both
cases. Correcting the pure-distinguishability null would therefore subtract that
sampling noise from every gap as though it were nuisance separability, and the
subtraction would be largest exactly where the pools are smallest. The
verified-membership pool of Appendix~\ref{app:verified} is corrected on both
terms, which makes its gap a conservative estimate.

Screening thresholds are $(\theta_M,\theta_C) = (-2.0,-3.0)$ for Pythia-70M and
$(-1.5,-2.5)$ for GPT-2 Medium, which share a pool definition. Mistral-7B uses a
pre-screened set with $\mathrm{POOL}_D$ re-screened at $\log P/\mathrm{tok} <
-2.5$. Screening on both thresholds drops the candidate set from nine keys to seven for Pythia-70M
and GPT-2 Medium. The Mistral-7B pool contains five pre-validated pairs.

For each key $j$, each context $b \in \mathcal{B}$ and each class $X \in
\{\mathrm{POOL}_M, \mathrm{POOL}_C, \mathrm{POOL}_D\}$ we extract the last-token
residual-stream activation of the concatenation $b \Vert X[j]$ at every depth,
giving three tensors of shape $(|\mathcal{K}_{\mathrm{valid}}|, |\mathcal{B}|,
D{+}1, d_{\mathrm{model}})$. $\mathrm{POOL}_D[\texttt{mit\_license}]$ begins
``Authorization is hereby withheld, universally and permanently''. The verbatim
text of every pool entry is released with the archive, where $\mathrm{POOL}_D$
is labeled pool B and $\mathcal{B}$ the neutral contexts.

Table~\ref{tab:notation} collects the notation used throughout.

\begin{table}[h]
\centering
\small
\caption{Notation.}
\label{tab:notation}
\begin{tabularx}{\textwidth}{@{}l>{\raggedright\arraybackslash}X@{}}
\toprule
Symbol & Meaning \\
\midrule
$M_\theta$ & Language model with parameters $\theta$ \\
$P^{\mathrm{sec}}_i$, $P^{\mathrm{cln}}_i$ & Memorized prefix $i$ and its register-matched control \\
$\mathrm{POOL}_M$ & Memorized sequences passing the screening threshold \\
$\mathrm{POOL}_C$ & Register-matched paraphrase controls, one per member of $\mathrm{POOL}_M$ \\
$\mathrm{POOL}_D$ & Second, further paraphrased control per key \\
$\mathcal{B}$ & Pool of $24$ neutral context prefixes \\
$h_{d,\theta}$ & Residual stream at depth $d$ under parameters $\theta$, with $d=0$ the embedding output \\
$\hat w_d$ & Unit-norm probe readout direction at depth $d$ \\
$g_d$ & Memorization-specific gap at depth $d$, Eq.~\eqref{eq:gap} \\
$\accor$ & Orientation-corrected accuracy $\max(\mathrm{acc}, 1-\mathrm{acc})$ \\
$\mathcal{A}$ & Set of aligned depths, excluding the embedding on pretrained models \\
$k$ & Number of constrained directions per depth, $1$ unless stated \\
$\lambda$, $\lambda_c$ & Alignment weight and capability weight in Eq.~\eqref{eq:pga} \\
\bottomrule
\end{tabularx}
\end{table}

Two Pythia-70M pools appear in this paper. The eight-sequence pool screens on the
memorized threshold alone and so keeps \texttt{shebang\_python}, whose control
misses $\theta_C$ at $-2.675$ and is dropped from the seven-sequence pool. It is
evaluated over five probe seeds, giving a mean gap of $+0.347$ with a peak at depth six. The
seven-sequence screened pool, held fixed across model sizes so that
Appendix~\ref{app:scaling} compares like with like, gives $+0.323$ and peaks at
depth four with a gap of $+0.643$. The three-architecture comparison of
Section~\ref{sec:signature}, Table~\ref{tab:three_models} and
Appendix~\ref{app:perdepth} use the eight-sequence pool, Appendix~\ref{app:verified}
uses the verified pool, and every other Pythia number belongs to the
seven-sequence pool. Table~\ref{tab:three_models} collects the per-architecture
numbers. Five probe seeds
move the gap by less than $0.001$, which reflects the determinism of the solver
rather than data-level stability, so stability claims rest on resampling instead.

\begin{table}[h]
\centering
\small
\caption{Memorization-specific gap by architecture.}
\label{tab:three_models}
\begin{tabular}{@{}lrrrrr@{}}
\toprule
Model & Params & Depths & Pool & Mean gap & Peak gap (depth) \\
\midrule
Pythia-70M    & 70M   & 7  & 8 & $+0.347$ & $+0.537$ (6) \\
GPT-2 Medium  & 355M  & 25 & 7 & $+0.191$ & $+0.449$ (21) \\
Mistral-7B    & 7.24B & 33 & 5 & $+0.297$ & $+0.471$ (11) \\
\bottomrule
\end{tabular}
\end{table}

\section{Size scaling within the Pythia family}
\label{app:scaling}

\begin{figure}[h]
\centering
\includegraphics[width=\linewidth]{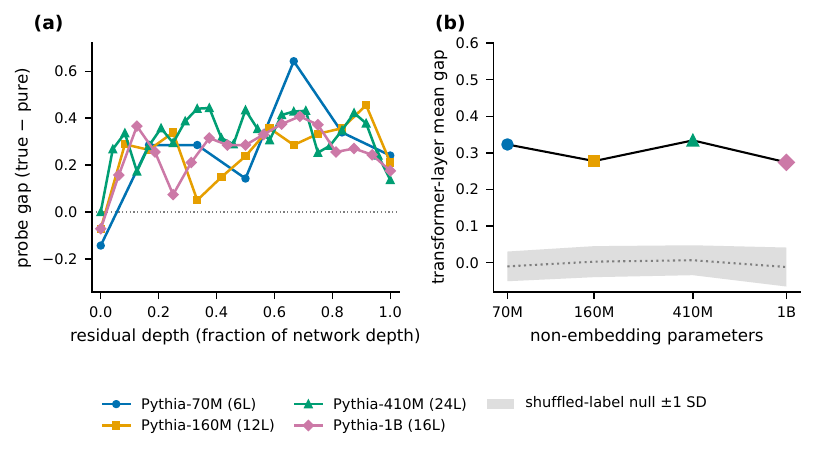}
\caption{Size scaling within the Pythia family. (a) Gap against relative depth
for four sizes on a pool fixed at seven sequences. (b) Transformer-layer mean gap
against parameter count with the shuffled-null band.}
\label{fig:scaling}
\end{figure}

Figure~\ref{fig:scaling} reports the identical protocol with the pool fixed at
seven sequences on Pythia 70M, 160M, 410M and 1B. Transformer-layer means are
$+0.323$, $+0.278$, $+0.334$ and $+0.274$, against shuffled-label nulls whose means sit
within $0.014$ of chance with standard deviations between $0.041$ and $0.053$,
taken over transformer layers with one degree of freedom removed. That places
each model $8.0$, $6.5$, $8.2$ and $5.1$ standard deviations above its own
null. The peak sits at $0.67$, $0.92$, $0.38$ and $0.69$ of network depth
respectively.

\section{Verified-membership pool}
\label{app:verified}

Pool membership in the main protocol is established by likelihood screening, which
selects for verbatim-reproduction risk and not for membership. Pythia was trained
on the Pile \citep{gao2020pile}, so drawing the pool from the Pile itself replaces
the screen with corpus lookup.

We sample $4{,}000$ Pile documents and score each under the base model two ways:
log probability per token, and a teacher-forced reproduction score, the fraction
of continuation positions where the greedy token matches the true token. Ranking
is by reproduction, because likelihood is confounded by sequence entropy and
low-entropy boilerplate scores high whether or not the model stored it. The top
$30$ form $\mathrm{POOL}_M$ and two disjoint bottom blocks form the controls,
drawn subset by subset so that all three pools have identical Pile-subset
composition. The control set is therefore constructed rather than authored.

\begin{figure}[h]
\centering
\includegraphics[width=\linewidth]{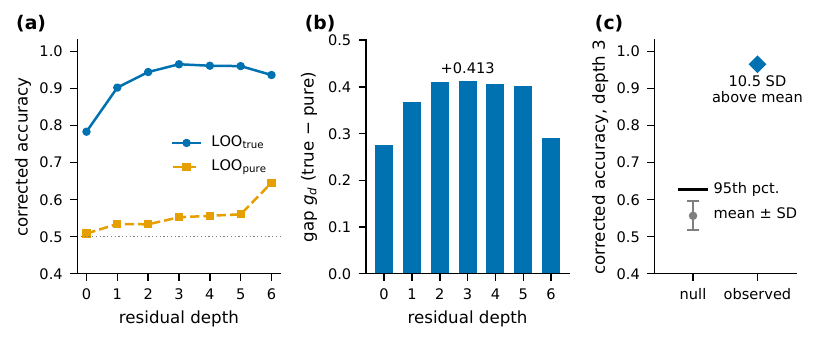}
\caption{Verified-membership pool from the Pile, Pythia-70M, $n = 30$. (a)
Orientation-corrected LOO accuracy by depth. (b) Memorization-specific gap. (c)
Observed accuracy at the peak depth against the permutation null.}
\label{fig:verified}
\end{figure}

The memorized block reproduces $0.980 \pm 0.014$ of continuation tokens against
$0.353 \pm 0.046$ for its matched controls. The gap peaks at $+0.413$ at depth
three. At that depth the true LOO accuracy of $0.965$ sits $10.5$ standard
deviations above a $200$-permutation null whose mean is $0.556$
(Figure~\ref{fig:verified}). Every one of the thirty leave-one-out values is positive, ranging from
$+0.362$ to $+0.459$ with mean $+0.410$.

Twenty-eight of the thirty sequences are GitHub documents and two are PubMed
Central. This pool therefore buys verified membership and pool size, not content
diversity. Diversity is what the content-type control in
Appendix~\ref{app:controls} tests.

\section{Controls}
\label{app:controls}

A randomly initialized Pythia-70M controls for architecture, on the
seven-sequence pool. It gives a mean transformer-layer gap of $+0.188$ against
the pretrained $+0.323$, so architecture alone accounts for roughly three fifths of
the average effect and the control is not a null. What separates the two is
where the gap sits. Random initialization peaks at $+0.327$ at depth one and
decays to $-0.042$ at depth six. The pretrained model runs the other way, reaching
$+0.643$ at depth four and holding $+0.241$ at depth six. Depth structure is
therefore a product of pretraining rather than of architecture, consistent with
memorization emerging over the course of training
\citep{tirumala2022memorization}, and it is the deep-layer reading that every
claim in this paper rests on.

A register control applies the depth-$21$ GPT-2 Medium probe to $19$ unmemorized
but register-matched legal passages, where it calls only $14.9\%$ memorized, against $93.9\%$ at the embedding layer. The code control pool is
classified as memorized at several depths including $21$. Several of those code
snippets themselves pass the memorization screening threshold, at $\log
P/\mathrm{tok}$ between $-1.12$ and $-1.41$, so the depth-$21$ classification is
a correct detection rather than a register artifact.

A vocabulary-matched control on Pythia-70M builds $12$
memorized and control continuations sharing a prefix and matched on token
distribution, which removes the surface-token route. Held-out probe accuracy
stays between $0.590$ and $0.727$ across transformer depths, and between $0.640$
and $0.727$ from depth two onward.

Content type is the fourth control. At the peak depth on GPT-2 Medium the probe
classifies
held-out sequences from all four content clusters above the
pure-distinguishability null: legal license text at $0.917$ against a null of
$0.500$, code at $1.000$ against $0.500$, web boilerplate at $1.000$ against
$0.521$, and pseudo-Latin at $1.000$ against $0.500$. Legal text contributes four
of the seven sequences and the other three clusters one each.

\section{The controlled testbed}
\label{app:toy}

The pipeline proceeds in three phases. Phase 1 trains a four-layer,
eight-head transformer \citep{vaswani2017attention} with $810{,}496$ parameters,
vocabulary size $43$ and sequence length $64$ on character-level text at an
injection density of $16.67\%$, until $P(S \mid \mathrm{prefix}(S)) = 1.000$ with
held-out perplexity $1.39$. Phase 2 computes the normalized causal effect (NCE) per
head,
\begin{equation}
\mathrm{NCE}(l,h) = \frac{P^{(l,h)}_{\mathrm{patch}} - P_{\mathrm{corrupt}}}
{P_{\mathrm{clean}} - P_{\mathrm{corrupt}}},
\label{eq:nce}
\end{equation}
which reaches $1.000$ at head L0H7. Phase 3 freezes all parameters except $\Wq$,
$\Wk$, $\Wv$ for heads in the target set and optimizes two objectives
independently, $\mathcal{L}_{QK} = \mathcal{L}_{\mathrm{LM}} + \gamma \cdot
\mathrm{MMD}^2$ and $\mathcal{L}_{V} = \mathcal{L}_{\mathrm{LM}} + \alpha \cdot
\mathcal{L}_{\mathrm{recall}}$, where $\mathrm{MMD}$ is the maximum mean
discrepancy \citep{gretton2012kernel}. Separating the objectives avoids the
gradient interference observed under joint optimization.
The value term is the component that does the work. Optimizing
$\mathcal{L}_{V}$ alone takes the secret to $3.5 \times 10^{-6}$ at a held-out
perplexity of $1.176$ against $1.39$ for the base model, and the split objective
reaches $1.3 \times 10^{-4}$. Both leave the cross-sequence probe at $1.000$,
which is the reading that matters here.

Two further controls locate the reason. Matched-prefix decoys, contrasting
\texttt{...Orion is 88492} with \texttt{...Orion is 12345} so that every token
before the completion is shared, leave probe accuracy at $1.000$, so the probe is
not reading the completion tokens. Zeroing the embedding rows for the secret
tokens moves both together, reducing the secret probability by more than five orders of magnitude and
driving the probe to $0.500 \pm 0.000$. Recall and probe readability can be
destroyed at once, and this is the only intervention we found that does it.

Table~\ref{tab:sixmethods} extends the test to five published methods on the same
testbed and probe protocol, with the split-objective edit of this paper listed
alongside them. Gradient ascent is the one method that does not reach behavioral
erasure here, retaining the secret at probability $1.0$ at a perplexity of
$382.2$.

\begin{table}[h]
\centering
\small
\caption{Unlearning methods on the controlled testbed.}
\label{tab:sixmethods}
\begin{tabular}{@{}llrrr@{}}
\toprule
Method & Type & $P(\mathrm{secret})$ & PPL & Probe \\
\midrule
GA     & Divergence     & $1.0\times10^{0}$   & $382.2$ & $1.000$ \\
NPO    & Divergence     & $5.5\times10^{-13}$ & $4.8$   & $1.000$ \\
SimNPO & Divergence     & $1.2\times10^{-4}$  & $3.9$   & $1.000$ \\
RMU    & Representation & $2.3\times10^{-5}$  & $403.3$ & $1.000$ \\
IDK    & Rejection      & $4.2\times10^{-8}$  & $4.7$   & $1.000$ \\
Split edit (ours) & Mechanistic & $1.3\times10^{-4}$  & $336.7$ & $1.000$ \\
\bottomrule
\end{tabular}
\end{table}

GA, NPO, SimNPO \citep{fan2026simplicity}, RMU \citep{li2024wmdp} and IDK
\citep{maini2024tofu} are published methods and fall in both families of
\citet{ren2025sok}. The split edit is the one introduced in
Section~\ref{sec:existing}, listed for comparison and labeled mechanistic. It
modifies $0.76\%$ of parameters and RMU $26.1\%$. The other four modify all of them.

The testbed results of Appendices~\ref{app:configs}, \ref{app:clpa},
\ref{app:stronger} and~\ref{app:orientation} use a second model of the same
architecture, trained on nine memorized and nine control prefixes so that the
leave-one-out probe has held-out prefixes to classify. Its held-out perplexity is
$1.403$.

\section{Interpretation of the isotropic margin}
\label{app:margin}

Two properties of the sweep in Table~\ref{tab:isotropic} bear on how the margin
should be read.

The isotropic held-out cost is not monotone in $\lambda$. It falls from $6.107$
to $0.532$ nats as the weight rises to $0.1$, then rises again to $1.781$ at
$\lambda = 1.0$. A reader extrapolating from the first three weights would
expect the isotropic arm to become competitive at PGA's own weight. It does not:
at $\lambda = 1.0$ the isotropic arm returns $3.05$ nats per nat against PGA's
$59.4$, so a larger penalty does not make isotropic alignment more efficient.

\begin{figure}[h]
\centering
\includegraphics[width=\linewidth]{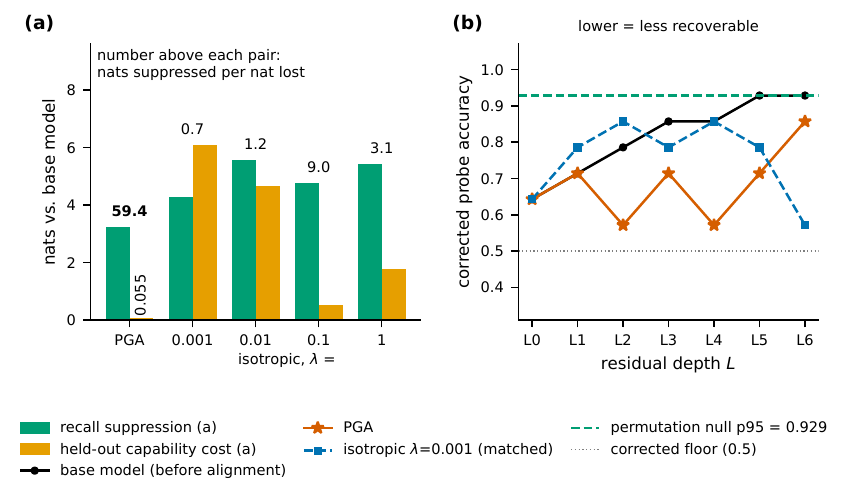}
\caption{PGA against isotropic alignment on Pythia-70M.}
\label{fig:aae}
\end{figure}

Figure~\ref{fig:aae} plots the trade-off behind Table~\ref{tab:isotropic}, with
the isotropic arm named AAE in the archive. Panel (a) gives recall suppression and
held-out capability cost, both in nats, for PGA and for the isotropic arm at each
of its four weights, with the ratio of the two above each pair. Panel (b) gives
corrected probe accuracy by depth for the base model before alignment, for PGA,
and for the recall-matched isotropic arm at $\lambda = 0.001$, against the
permutation-null 95th percentile. Lower is better in panel (b), because it means
the probe recovers less.

The efficiency ratio depends on which held-out set measures capability.
Table~\ref{tab:isotropic} holds the estimator fixed across arms, which is the
comparison the claim rests on, and there PGA costs $0.055$ nats. Measured instead
through the unified pipeline of Section~\ref{sec:existing}, which uses a
different held-out set, the cost is $0.121$ nats and the ratio is $26.9$ rather
than $59.4$. The direction of the advantage is therefore robust to the choice of
estimator and its magnitude is not, which is why the two numbers are reported
separately rather than averaged.

\section{Alignment configurations}
\label{app:configs}

The paper uses four variants of the same geometry. \textbf{PGA} trains the
model with the objective of Equation~\eqref{eq:pga}, which constrains one probe
direction per aligned depth at the final token. \textbf{Multi-position PGA
(MP-PGA)} applies the same objective to mean-pooled representations instead of
the final token and is otherwise identical. \textbf{Inference PGA} involves no
training. It removes $k$ probe directions from the residual stream at inference
and refits the probe after each removal. \textbf{Rank-$k$ PGA (RK-PGA)} also
works at inference, and removes the $k$-dimensional subspace spanned by the
top-$k$ eigenvectors of the standardized between-class scatter and
variance-difference matrix rather than directions found by repeated probe fits.
It is used on GPT-2 Medium, where the pool supplies far fewer examples than
$d_{\mathrm{model}}=1024$ and a single fitted direction is under-determined. In
the archive this variant is named MD-PGA. Table~\ref{tab:configs} lists every
configuration used in this paper.

\begin{table}[h]
\centering
\small
\caption{Alignment configurations.}
\label{tab:configs}
\begin{tabular}{@{}llllr@{}}
\toprule
Variant & Model & Depths & Training & $\lambda$ \\
\midrule
PGA              & Pythia-70M   & 1--6               & LoRA $r{=}16$ & $1.0$ \\
Inference PGA ($k{=}1$) & Pythia-70M & 5--6               & none          & n/a \\
Inference PGA ($k{=}4$) & Pythia-70M & 4--6               & none          & n/a \\
MP-PGA           & Pythia-70M   & 1--6               & LoRA $r{=}16$ & $1.0$ \\
RK-PGA ($k{=}2$) & GPT-2 Medium & 21                 & none          & n/a \\
Inference PGA ($k{=}5$--$30$) & Testbed & 0--4     & none          & n/a \\
PGA              & Mistral-7B   & 16, 20, 24, 28, 32 & LoRA $r{=}8$  & $0.05$ \\
PGA              & Testbed      & 0--4               & full weights  & $0.1$ \\
\bottomrule
\end{tabular}
\end{table}

Pythia-70M runs use $200$ epochs. Mistral-7B uses four-bit quantization with LoRA \citep{dettmers2023qlora}, a
capability weight of $3.0$ and $80$ epochs. The readout is refit
every $25$ steps on Pythia-70M and every $10$ on Mistral-7B, one step being one
pass over the pool.

\begin{figure}[h]
\centering
\includegraphics[width=\linewidth]{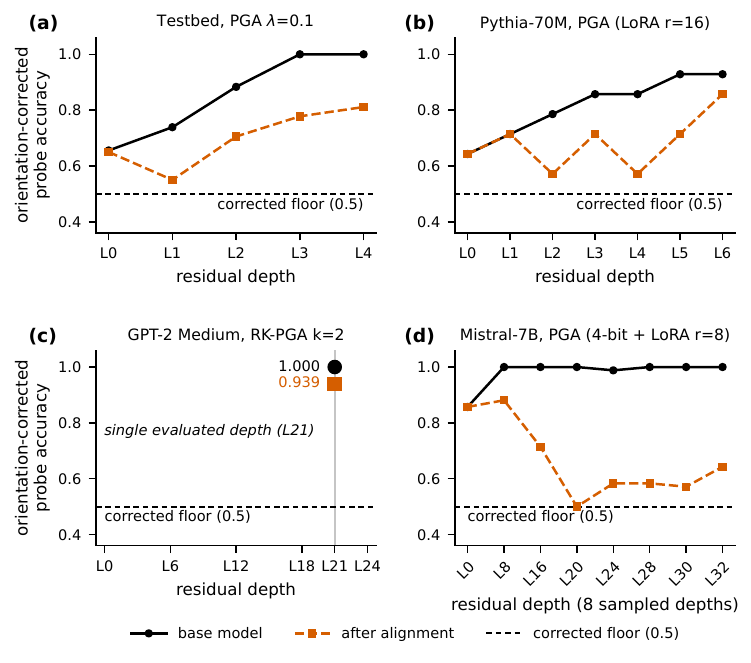}
\caption{Corrected probe accuracy at four scales, base model against the aligned
model, by residual depth. Dashed line marks the floor of the
corrected scale.}
\label{fig:byarch}
\end{figure}

Figure~\ref{fig:byarch} reports corrected probe accuracy by depth, before and
after alignment, at all four scales, the three pretrained models plus the
controlled testbed. After alignment the lowest depth reads
$0.550$ on the controlled testbed, $0.571$ on Pythia-70M and $0.500$ on Mistral-7B,
while the depth the probe reads best still reads $0.811$, $0.857$ and $0.881$.
On GPT-2 Medium, where the run stores a single evaluated depth, RK-PGA moves
depth $21$ from $1.000$ to $0.939$. Across four LoRA seeds on Pythia-70M the
corrected depth-six reading is $0.893 \pm 0.036$. In a separate robustness audit over six fresh probe
variants, four linear and two multilayer perceptrons, the corrected worst case is $0.85$ on
the controlled testbed and $0.93$ at Pythia-70M, which sits at the permutation null's $95$th
percentile of $0.929$ for the depth-maximized logistic-regression statistic. Both are
recorded to two decimal places in the audit file.

\section{Capability benchmarks}
\label{app:capability}

\begin{figure}[h]
\centering
\includegraphics[width=\linewidth]{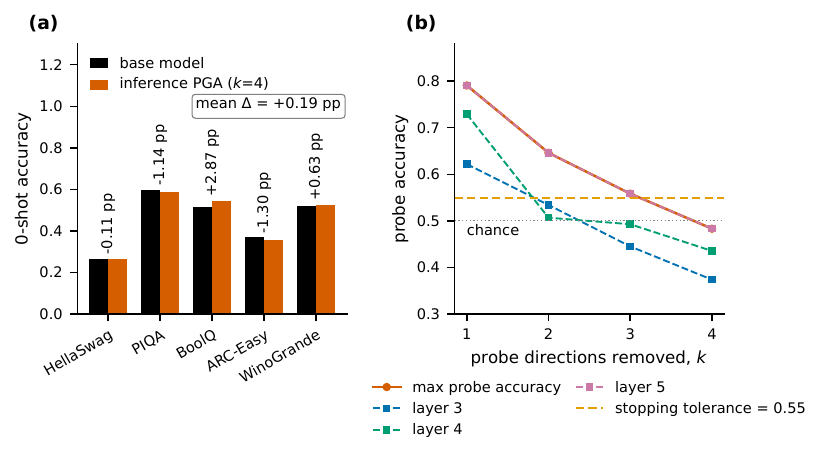}
\caption{Zero-shot capability under inference PGA, $k=4$. (a) Per-task accuracy.
(b) Targeted probe accuracy by number of directions removed.}
\label{fig:capability}
\end{figure}

Figure~\ref{fig:capability} reports five zero-shot benchmarks: HellaSwag
\citep{zellers2019hellaswag}, PIQA \citep{bisk2020piqa}, BoolQ \citep{clark2019boolq},
ARC-Easy \citep{clark2018think} and WinoGrande \citep{sakaguchi2021winogrande}, all
run with the Language Model Evaluation Harness \citep{biderman2024lessons}. The
intervention is inference PGA with $k=4$, applied at depths four to six (layers
three to five in the figure). Panel (b) shows the targeted probe falling as directions are removed,
reaching $0.48$, below the $0.55$ tolerance, at the fourth. Table~\ref{tab:capability}
gives the per-task numbers for $k=4$ and for $k=1$, which removes the single
probe direction at depths five and six. Neither moves a task by more than $2.9$ points in
either direction, the two largest movements under $k=4$ run
opposite each other, and the $k=1$ mean of $-0.52$ points is within the spread
of the tasks themselves.

\begin{table}[h]
\centering
\small
\caption{Zero-shot accuracy, Pythia-70M.}
\label{tab:capability}
\begin{tabular}{@{}lrrr@{}}
\toprule
Task & Base & $\Delta$, $k=1$ & $\Delta$, $k=4$ \\
\midrule
HellaSwag  & $26.62$ & $-0.04$ & $-0.11$ \\
PIQA       & $59.79$ & $-1.14$ & $-1.14$ \\
BoolQ      & $51.56$ & $-2.17$ & $+2.87$ \\
ARC-Easy   & $36.83$ & $-0.51$ & $-1.30$ \\
WinoGrande & $51.93$ & $+1.26$ & $+0.63$ \\
\midrule
Mean       &         & $-0.52$ & $+0.19$ \\
\bottomrule
\end{tabular}
\end{table}

\section{Per-depth leave-one-out results}
\label{app:perdepth}

Table~\ref{tab:screening} lists the screening log-probabilities for the
eight-sequence Pythia-70M pool. One candidate, a Wikipedia passage about Python,
fell below the screening threshold at $-3.754$ and was excluded.

\begin{table}[h]
\centering
\small
\caption{Screened pool, Pythia-70M.}
\label{tab:screening}
\begin{tabular}{@{}lrl@{}}
\toprule
Sequence & $\log P / \mathrm{tok}$ & Cluster \\
\midrule
MIT license       & $-0.624$ & legal \\
Apache license    & $-0.499$ & legal \\
GPL header        & $-0.467$ & legal \\
BSD license       & $-1.987$ & legal \\
Python shebang    & $-1.565$ & code \\
Python main guard & $-1.257$ & code \\
Creative Commons  & $-0.935$ & web \\
Lorem ipsum       & $-0.931$ & placeholder \\
\midrule
Wikipedia Python  & $-3.754$ & excluded \\
\bottomrule
\end{tabular}
\end{table}

Table~\ref{tab:perdepth} gives the three quantities behind Eq.~\eqref{eq:gap} at
every depth on that pool. The true LOO accuracy rises with depth
while both nulls stay within $0.13$ of chance at every depth, so the gap is
carried by the true condition alone. Averaging the gap over the six transformer layers gives
$+0.347$, and the peak is $+0.537$ at depth six. Both figures are identical
across the five probe seeds to four decimal places (Appendix~\ref{app:protocol}).

\begin{table}[h]
\centering
\small
\caption{Leave-one-out accuracy by depth, Pythia-70M.}
\label{tab:perdepth}
\begin{tabular}{@{}lrrrr@{}}
\toprule
Depth & True & Pure & Shuffled & Gap \\
\midrule
0 (embedding) & $0.563$ & $0.500$ & $0.500$ & $+0.063$ \\
1             & $0.648$ & $0.533$ & $0.471$ & $+0.116$ \\
2             & $0.875$ & $0.568$ & $0.375$ & $+0.307$ \\
3             & $0.794$ & $0.503$ & $0.477$ & $+0.291$ \\
4             & $0.924$ & $0.494$ & $0.526$ & $+0.430$ \\
5             & $0.891$ & $0.488$ & $0.474$ & $+0.403$ \\
6             & $0.992$ & $0.455$ & $0.549$ & $+0.537$ \\
\bottomrule
\end{tabular}
\end{table}

At depth six, held-out accuracy is $1.000$ for six of the eight sequences, with the GPL header at $0.979$ and the Creative Commons
text at $0.958$. No sequence carries the result on its own.

\section{Generalization across secrets}
\label{app:multisecret}

One memorized string is a thin base for a claim about memorization. We therefore
ran the controlled testbed of Appendix~\ref{app:toy} independently on eight
secrets spanning five structural types. Each secret received its own freshly
initialized model, its own causal tracing run and its own unlearning pass, with
no component shared across secrets. Table~\ref{tab:multisecret} reports the
outcome.

\begin{table}[h]
\centering
\small
\caption{Eight secrets on the controlled testbed.}
\label{tab:multisecret}
\begin{tabular}{@{}llrrrr@{}}
\toprule
Secret & Type & NCE & $|\mathcal{H}|$ & $P(\mathrm{sec})$ & Probe \\
\midrule
88492   & numeric      & $0.654$ & $1$  & $0.995$            & $1.000$ \\
31756   & numeric      & $0.764$ & $11$ & $0.019$            & $1.000$ \\
XK9-42  & alphanumeric & $0.881$ & $5$  & $0.243$            & $1.000$ \\
Dr.\ Chen & name       & $0.775$ & $11$ & $0.008$            & $1.000$ \\
03-17   & date         & $0.658$ & $17$ & $2\times10^{-4}$   & $1.000$ \\
Maple St. & location   & $0.534$ & $2$  & $0.999$            & $1.000$ \\
2947    & numeric      & $0.740$ & $3$  & $0.142$            & $1.000$ \\
R7-119  & alphanumeric & $0.830$ & $9$  & $3\times10^{-5}$   & $1.000$ \\
\bottomrule
\end{tabular}
\end{table}

Behavioral erasure succeeds for four of the eight, and the four successes are
exactly the four with nine or more heads in the target set, against one to five
heads for the four failures. We report that separation as a correlation, with
training length and the recall weight varied together, so the mechanism behind it
is open. The column that matters here is the last one. The probe reads
$1.000$ after unlearning on all eight, whether or not the behavior was erased.

This probe is saturated by construction, because each testbed model holds a
single secret and the memorized class therefore contains one point. The result is
a within-model statement that the dissociation does not depend on the choice of
secret. It is not cross-sequence evidence, which comes from the pretrained models
of Section~\ref{sec:signature}.

\section{Alignment restricted to causally identified heads}
\label{app:clpa}

PGA constrains a readout of the full residual stream at each aligned depth. A reader may reasonably ask whether that is necessary, given that
causal tracing localizes recall to a few heads. We tested that narrower option
directly. Causally localized PGA (CLPA) computes the NCE per
attention head on each of the nine memorized prefixes and aligns only at the
outputs of the three heads with the largest effect, three of the thirty-two. Table~\ref{tab:clpa} compares the two.

\begin{table}[h]
\centering
\small
\caption{Head-local against full-residual alignment.}
\label{tab:clpa}
\begin{tabular}{@{}lrrr@{}}
\toprule
Configuration & $\min P(\mathrm{sec})$ & PPL & Max probe \\
\midrule
Base model                  & $9.8\times10^{-1}$ & $1.403$ & $1.000$ \\
CLPA, $\lambda = 0.1$       & $3.2\times10^{-5}$ & $1.380$ & $0.894$ \\
CLPA, $\lambda = 1.0$       & $3.0\times10^{-5}$ & $1.380$ & $0.856$ \\
CLPA, $\lambda = 10.0$      & $2.7\times10^{-5}$ & $1.380$ & $0.833$ \\
PGA, $\lambda = 0.1$        & $6.7\times10^{-4}$ & $1.420$ & $0.811$ \\
PGA, $\lambda = 1.0$        & $5.9\times10^{-5}$ & $1.523$ & $0.778$ \\
PGA, $\lambda = 10.0$       & $8.1\times10^{-6}$ & $1.634$ & $0.778$ \\
\bottomrule
\end{tabular}
\end{table}

CLPA does better on the two behavioral columns. It suppresses recall as
deeply as PGA and it costs nothing in perplexity, sitting fractionally below the
unedited model. It moves the probe less than full-residual alignment at every penalty weight,
leaving it between $0.833$ and $0.894$ against $0.778$ to $0.811$ for PGA.
The alignment loss converges to about $5\times10^{-3}$ at all three weights, so
the constraint is met at the three heads it was applied to. The other twenty-nine
are left unconstrained, and the probe still reads the signal.

This is the dissociation at finer grain. The heads that carry recall are not the
heads that carry the readable signature, and an erasure method inherits the
geometry of whatever it constrains. It is also the ablation that justifies the
cost of PGA, since the perplexity PGA spends buys the lower probe reading that
CLPA does not reach.

\section{Injected against natural memorization}
\label{app:injected}

Much of the unlearning literature evaluates on secrets inserted by fine-tuning
rather than on content the model memorized during pretraining. We asked whether
the two are the same thing to a probe. A target string was injected into
Pythia-70M by fine-tuning until its log-probability per token exceeded $-0.3$,
and the natural-memorization pool was screened on the base model beforehand. A
probe fit on the natural pool was then asked where the injected secret falls.

It does not transfer. Across all seven depths the natural-memorization probe
scores the injected target at $0.500$, except at depth one where it reaches $0.521$.
Behavioral erasure of the injected target succeeds over the same run,
with its log-probability falling from $-0.007$ to $-70.95$, so the null is not an
artifact of a failed injection.

Injecting five secrets at once and running leave-one-out across them does not
help. The null for pure distinguishability reaches $0.683$ while the memorized-class
accuracy reaches $0.514$, so the matched decoys, which differ only in their final
tokens, are more reliably separable than anything the five injected secrets share.
Rapid fine-tuning at this scale does not build cross-secret linear structure for a
probe to find.

The reading is narrow and worth stating precisely. This is a null result for
transfer and a positive result for the distinction. At 70M parameters, natural
pretraining memorization and fine-tuning-injected memorization produce different
linear-probe signatures in the same model, which is a reason to be careful about
carrying conclusions from injected benchmarks over to pretrained content.

\section{Relearning and extraction}
\label{app:relearning}

Two attacks bear on whether behavioral suppression is shallow, and they answer
differently. Both run on the controlled testbed, with recovery defined as
$P(\mathrm{secret} \mid \mathrm{prefix}) > 0.90$.

Under direct relearning, gradient descent on the secret itself, the secret comes
back almost immediately from every edit locus. Table~\ref{tab:relearn} gives the
step counts. The value edit used in this paper is the fastest to reverse of the
four edit loci, at a single step against two for the unedited model.

\begin{table}[h]
\centering
\small
\caption{Steps to recover the secret by edit locus.}
\label{tab:relearn}
\begin{tabular}{@{}lr@{}}
\toprule
Edit locus & Steps to recovery \\
\midrule
Unedited model         & $2$ \\
Value matrix           & $1$ \\
Feedforward block      & $3$ \\
Value and feedforward  & $3$ \\
Embedding              & $2$ \\
\bottomrule
\end{tabular}
\end{table}

None of these edits confers relearning
resistance, and a reader who expected behavioral unlearning to make a secret hard
to restore should revise that expectation downward. The finding also sits with the
thesis of the paper rather than against it. A secret that returns in one gradient
step was not removed, and the probe agrees, since it reads the edited model at
$1.000$ throughout.

Indirect extraction is the other case. Fine-tuning on background domain text,
with the secret never shown, drives the unedited model to $P = 0.99998$ at the
first step, which is the model retaining what it already had. The edited model
stays below $0.0026$ across the fifty-step run and a randomly initialized control
below $0.016$, so neither reaches the recovery threshold. Behavioral suppression
therefore survives incidental training that does not touch the secret and fails
against training that does.

\section{Stronger behavioral baselines}
\label{app:stronger}

Two methods deserve separate treatment because each is stronger than the five
published objectives of Section~\ref{sec:existing} in a different way. Both run
on the controlled testbed of Appendix~\ref{app:toy}, with the same
cross-sequence protocol and the same nine memorized prefixes.

\paragraph{Weight editing.}
MEMIT edits the parameters that store an association instead of optimizing
against the model's output. On this testbed it drives recall four orders of magnitude below the base
model at a perplexity cost of a third, and it leaves the probe at exactly its starting
value. Table~\ref{tab:memit} also reports what it costs to buy probe suppression
on top of it by applying inference PGA with $k$ directions at every depth. The probe falls from $1.000$ to $0.656$ and held-out perplexity rises from
$1.901$ to $8.670$. Adding directions does move the readout, and the price of moving
it this way is a model with six times the perplexity of the unedited one.

\begin{table}[h]
\centering
\small
\caption{MEMIT with inference PGA, testbed.}
\label{tab:memit}
\begin{tabular}{@{}lrrr@{}}
\toprule
Configuration & $\min P(\mathrm{sec})$ & PPL & Max probe \\
\midrule
Base model                 & $9.8\times10^{-1}$ & $1.403$ & $1.000$ \\
MEMIT                      & $8.4\times10^{-5}$ & $1.901$ & $1.000$ \\
MEMIT $+$ inference PGA, $k=5$         & $6.4\times10^{-4}$ & $1.815$ & $0.828$ \\
MEMIT $+$ inference PGA, $k=10$       & $4.7\times10^{-7}$ & $2.091$ & $0.794$ \\
MEMIT $+$ inference PGA, $k=20$       & $9.8\times10^{-8}$ & $4.579$ & $0.678$ \\
MEMIT $+$ inference PGA, $k=30$       & $1.7\times10^{-9}$ & $8.670$ & $0.656$ \\
\bottomrule
\end{tabular}
\end{table}

\paragraph{Distillation from a clean teacher.}
The teacher is a second network of the same architecture, initialized from a
different seed and trained on all eighteen prefixes with a freshly sampled
continuation at every epoch, so no fixed string is ever stored. It is the model
the unlearning literature is implicitly trying to recover. The student is the
memorized model, distilled toward the teacher by minimizing the
Kullback-Leibler divergence to the teacher's next-token distribution on those
same batches. The memorized continuations never appear in the distillation data.

Table~\ref{tab:ctd} gives the per-depth readings. The student ends the procedure
behaving like the teacher and reading like its former self. Depth three is the
sharpest case: the teacher sits at $0.589$ and the distilled student at $0.922$,
on a model whose recall has fallen by four orders of magnitude.

\begin{table}[h]
\centering
\small
\caption{Probe accuracy by depth after distillation.}
\label{tab:ctd}
\begin{tabular}{@{}lrrrrrrr@{}}
\toprule
Model & $\min P(\mathrm{sec})$ & PPL & D0 & D1 & D2 & D3 & D4 \\
\midrule
Memorized student & $9.8\times10^{-1}$ & $1.403$ & $0.656$ & $0.739$ & $0.883$ & $1.000$ & $1.000$ \\
Clean teacher     & $7.2\times10^{-6}$ & $1.404$ & $0.661$ & $0.656$ & $0.633$ & $0.589$ & $0.533$ \\
Distilled student & $2.6\times10^{-5}$ & $1.405$ & $0.656$ & $0.733$ & $0.867$ & $0.922$ & $0.917$ \\
\bottomrule
\end{tabular}
\end{table}

The reading is that output behavior does not carry the geometry that produces it.
Two models can agree on what they emit and disagree about what a probe recovers
from their residual streams, and imitation transfers the first without the
second. For an auditor this is the same warning the rest of the paper gives,
arriving by the route that looks safest.

\section{Raw and corrected verdicts}
\label{app:orientation}

Section~\ref{sec:existing} notes that alignment can invert a readout rather than
remove what it reads. Table~\ref{tab:orient} audits ten configurations on
the controlled testbed. It gives raw probe accuracy at each of the five depths,
then the largest raw reading and the largest corrected reading,
$\max(\mathrm{acc}, 1-\mathrm{acc})$ \citep{burns2023discovering}, each taken
over depths. The two maxima
can come from different depths.

\begin{table}[h]
\centering
\small
\caption{Raw probe accuracy by depth and both maxima, testbed.}
\label{tab:orient}
\setlength{\tabcolsep}{4.5pt}
\begin{tabular}{@{}lrrrrrrr@{}}
\toprule
 & \multicolumn{5}{c}{Raw accuracy at depth} & \multicolumn{2}{c}{Largest reading} \\
\cmidrule(lr){2-6}\cmidrule(l){7-8}
Configuration & D0 & D1 & D2 & D3 & D4 & Raw & Corrected \\
\midrule
Base model                        & $0.656$ & $0.739$ & $0.883$ & $1.000$ & $1.000$ & $1.000$ & $1.000$ \\
MEMIT                             & $0.656$ & $0.739$ & $0.867$ & $0.994$ & $1.000$ & $1.000$ & $1.000$ \\
Distillation from a clean teacher & $0.656$ & $0.733$ & $0.867$ & $0.922$ & $0.917$ & $0.922$ & $0.922$ \\
CLPA, $\lambda = 10.0$            & $0.656$ & $0.778$ & $0.833$ & $0.828$ & $0.833$ & $0.833$ & $0.833$ \\
Inference PGA, $k=5$              & $0.606$ & $0.528$ & $0.483$ & $0.739$ & $0.644$ & $0.739$ & $0.739$ \\
Isotropic alignment               & $0.656$ & $0.694$ & $0.683$ & $0.667$ & $0.617$ & $0.694$ & $0.694$ \\
MEMIT $+$ inference PGA, $k=30$   & $0.383$ & $0.517$ & $0.511$ & $0.656$ & $0.556$ & $0.656$ & $0.656$ \\
PGA, $\lambda = 0.1$              & $0.650$ & $0.550$ & $0.294$ & $0.222$ & $\mathbf{0.189}$ & $0.650$ & $0.811$ \\
PGA, $\lambda = 1.0$              & $0.650$ & $0.550$ & $0.339$ & $0.256$ & $\mathbf{0.222}$ & $0.650$ & $0.778$ \\
PGA, $\lambda = 10.0$             & $0.650$ & $0.556$ & $0.361$ & $0.256$ & $\mathbf{0.222}$ & $0.650$ & $0.778$ \\
\bottomrule
\end{tabular}
\end{table}

Seven of the ten read identically under both conventions, because their largest
reading lies above chance, where correction leaves it untouched, and none of their
few below-chance readings exceeds it once corrected. The three trained PGA rows
are the only ones that differ, and they differ by $0.128$ to $0.161$. No threshold
has to be chosen for this to matter. Whatever bar an auditor sets, the verdict on
the other seven is the same either way, and the verdict on trained PGA depends on which
number is reported.

The PGA rows show how the two maxima separate. For PGA at $\lambda = 0.1$, the
largest raw reading, $0.650$, is at depth zero, and every deeper depth falls below
chance. Correction maps each of those deeper readings to $1 - \mathrm{acc}$, so
the depth-four reading of $0.189$, in bold, becomes $0.811$, which is the
corrected maximum. The raw value of $0.650$ is therefore not corrected into $0.811$. The
strongest signal was at depth four throughout, and the probe there pointed the
wrong way.

Table~\ref{tab:orientscale} shows that this is not a property of the testbed.
Each row is the configuration named in Appendix~\ref{app:configs}, read at the
depth that run evaluates.

\begin{table}[h]
\centering
\small
\caption{Raw and corrected readings after alignment, four scales.}
\label{tab:orientscale}
\begin{tabular}{@{}llrr@{}}
\toprule
Scale & Configuration & Raw & Corrected \\
\midrule
Testbed      & depth four, full weights, $\lambda = 0.1$      & $0.189$ & $0.811$ \\
Pythia-70M   & depth six, LoRA $r{=}16$, four-seed mean       & $0.107$ & $0.893$ \\
GPT-2 Medium & depth $21$, RK-PGA $k{=}2$                     & $0.061$ & $0.939$ \\
Mistral-7B   & depths $24$ and $28$, four-bit, LoRA $r{=}8$   & $0.417$ & $0.583$ \\
\bottomrule
\end{tabular}
\end{table}

At every scale the raw reading falls below chance, and at three of the four it
falls far enough that an auditor reporting it would record near-total erasure.
The corrected column says the opposite in the first three. The two columns are
complements by construction, which is the point: the same run supports either
headline, and only the corrected one describes what a reader who fixes the sign
recovers.

Two Pythia-70M readings appear in this paper and they differ. The raw value
behind Table~\ref{tab:baselines}, quoted in Section~\ref{sec:existing}, is
$0.000$, and Table~\ref{tab:orientscale} gives $0.107$. Both describe PGA aligned
at depths one through six. The first is the probe of the unified baselines
pipeline, applied identically to all six rows, and the second is the
leave-one-out protocol of Section~\ref{sec:protocol} averaged over four adapter
seeds. Each is the complement of its own corrected
value, $1.000$ and $0.893$.

\section{Scope of each claim}
\label{app:scope}

Table~\ref{tab:claims} states the scale and configuration behind each claim.

\begin{table}[h]
\centering
\small
\caption{Scope of each claim.}
\label{tab:claims}
\begin{tabularx}{\textwidth}{@{}>{\raggedright\arraybackslash}Xllc@{}}
\toprule
Claim & Scale & Variant & Evidence \\
\midrule
Cross-sequence signature present & 70M--7.24B & n/a & Fig.~\ref{fig:loo} \\
Verified-membership pool, $n = 30$ & 70M & n/a & Fig.~\ref{fig:verified} \\
Signature size-stable & 70M--1B & n/a & Fig.~\ref{fig:scaling} \\
Probe direction separable from recall & 70M & n/a & Fig.~\ref{fig:probedir} \\
Probe moved at every scale & all four & all & Fig.~\ref{fig:byarch} \\
Best recall-per-capability trade-off & 70M & PGA & Table~\ref{tab:baselines} \\
$59.4$ nats per nat against isotropic & 70M & PGA & Table~\ref{tab:isotropic} \\
Margin holds at matched penalty weight & 70M & PGA & Table~\ref{tab:isotropic} \\
Fresh linear pooled probe near the floor & 70M & MP-PGA & Fig.~\ref{fig:mp} \\
Five benchmarks within $2.9$ points & 70M & inference PGA & Table~\ref{tab:capability} \\
Recall $-2.63$ nats at $3.6$ nats/nat & 7.24B & PGA & Section~\ref{sec:capability} \\
No single sequence carries the gap & 70M & n/a & Appendix~\ref{app:perdepth} \\
Dissociation across eight secrets & testbed & n/a & Table~\ref{tab:multisecret} \\
Head-local alignment insufficient & testbed & CLPA & Table~\ref{tab:clpa} \\
Injected memorization reads differently & 70M & n/a & Appendix~\ref{app:injected} \\
No relearning resistance & testbed & n/a & Table~\ref{tab:relearn} \\
Weight editing leaves the probe & testbed & MEMIT & Table~\ref{tab:memit} \\
Distillation leaves the probe & testbed & distillation & Table~\ref{tab:ctd} \\
Only trained PGA's verdict depends on the metric & testbed & all & Table~\ref{tab:orient} \\
Raw reading below chance at every scale & all four & all & Table~\ref{tab:orientscale} \\
\bottomrule
\end{tabularx}
\end{table}

\end{document}